\documentclass{article}
\usepackage{graphicx}   
\usepackage{underscore}
\usepackage{listings}
\usepackage{xcolor} 
\definecolor{lightgray}{gray}{0.95}

\usepackage{hyperref}
\hypersetup{pdftex,colorlinks=true,allcolors=blue,pdfpagemode=UseOutlines}

\begin{document}
\title{Predicting Customer Lifetime Value in Free-to-Play Games}
\author{Paolo Burelli \\ IT University of Copenhagen - Tactile Games \\ pabu@itu.dk}
\date{}
\maketitle

\begin{abstract}
As game companies increasingly embrace a service-oriented business model, the need for predictive models of players becomes more pressing. Multiple activities, such as user acquisition, live game operations or game design need to be supported with information about the choices made by the players and the choices they could make in the future. 
This is especially true in the context of free-to-play games, where the absence of a pay wall and the erratic nature of the players' playing and spending behavior make predictions about the revenue and allocation of budget and resources extremely challenging.

In this chapter we will present an overview of customer lifetime value modeling across different fields, we will introduce the challenges specific to free-to-play games across different platforms and genres and we will discuss the state-of-the-art solutions with practical examples and references to existing implementations. 
\end{abstract}

\section{Introduction}

Customer lifetime value (CLV or LTV) refers broadly to the revenue that a company can attribute to one or more customer over the length of their relationship with the company~\cite{Pfeifer2004}. 
The process of predicting the lifetime value consists in producing one or more monetary values that correspond to the sum of all the different types of revenues that a specific customer, or a specific cohort, will generate in the future.
The purposes of this prediction are manifold: for example, having an early estimation of a customer's potential value allows more accurate budgeting for future investment; moreover, monitoring the remaining potential revenue from an established customer could permit preemptive actions in case of decreased engagement.

Predicting customer lifetime value is a complex challenge and, to date, there is no single established practice. Furthermore, due to its wide potential impact in different business aspects, the problem is being researched in different communities using a plethora of different techniques, varying from parametric statistical models to deep learning~\cite{Gupta2006a,Sifa2018}.
One of the initial primary drivers of research in CLV was direct marketing~\cite{Berger1998} with a focus on its use in marketing decision problems, such as acquisition or the trade-off between acquisition and retention costs~\cite{Blattberg1996}.
Research expanded progressively to other fields of marketing and customer relationship management, especially thanks to the pervasiveness of digital technology and the possibility to have more accurate tracking of the customer relationship~\cite{Jain2002}.

The increased quality and quantity of customer data available for analysis widened the areas of application of CLV related principles, and  helped accelerating the development of new models for CLV prediction.
Initial analytical models were developed on assumptions about constant and uniform characteristics of the user behavior (e.g. transaction frequency, margin of profit). Over time, models have evolved to take into account uncertainty and variations of user behavior~\cite{Schmittlein1987} and to be able to draw information from a wide spectrum of variables~\cite{Sifa2015}.

Models have traditionally been based, on transactional data, meaning that the variables describing the user behavior are based on different characteristics of the customers transactions with the company (e.g. recency, frequency, monetary (RFM)~\cite{Schmittlein1987}). However, with the advent of e-commerce and similar technologies, the available information about the customer behavior became increasingly richer, since it started to be possible to track not only the purchases, but also which objects were observed, how often would the customer visit the store and many other non-strictly transactional details.

One of the application area with probably the richest amount of user data is video games; however, until recently, due to its distribution and revenue models, the need for customer lifetime value prediction was relatively minor.
Traditionally, computer games distribution has relied on a premium pricing model, in which a game is developed, released and sold for a price. Following this pricing scheme, the monetization of the customer happens before the player starts playing the game and its post-purchase behavior does not affect the customer lifetime value directly~\cite{Baden-Fuller2013}.

Two main aspects radically changed the relationship between the game developers and the players/customers: digital distribution platforms and the free-to-play business model. 
Digital distribution platforms, such as Steam\footnote{Valve Corporation. Steam. https://store.steampowered.com 2018.} or the App Store\footnote{Apple Incorporated. App Store. https://www.apple.com/lae/ios/app-store 2018}, along with game developers' own on-line services allow tracking of user behavior beyond a single game, making it more meaningful and necessary to predict the customer lifetime value across multiple games, either being purchased on the same store or being linked to the same on-line account.

Free-to-play (F2P/Freemium) games follow a different revenue model than the traditional one: games are freely available, at no cost, to the players and the revenue comes from in-game advertisement and the sale of in-game items. These two revenue streams, contrarily to the classic game revenue model, begin only after the players/customers have been acquired and vary dramatically between players.
This kind of relationship between the company and the players resembles in part the relationship between a store and its customer, e.g. monetary transactions happen at irregular intervals, the transaction value is not constant and there is no explicit customer churn event.

However, differently from physical or on-line stores, the users are not only customers but also players, and the data available about their behavior includes a wide range of features that goes beyond transactions or goods browsing, such as players progression within the game or players' skill level.
Within the field of game analytics, the analysis of these features plays a major role in evaluating the quality and the potential success of a game~\cite{Drachen2013}; furthermore, the behavior has been linked with major business related metrics such as retention~\cite{Perianez2016} or customer lifetime value~\cite{Sifa2018}.

This abundance of accurate and high frequency data, covering different aspects of the customer experience make free-to-play-games an ideal application area for research in user modeling and prediction of future customer behavior. 
In turn, the development of such models would see an immediate application in the industry, as they would allow a better optimization of important processes such as user acquisition and customer relationship management~\cite{Lange2014}.

This, as well as the ever-expanding share of the game industry that leverages this business model, highlights customer life time value prediction as a key challenge in games data science and games research.
For these reasons, in this chapter, we will make an attempt to give an overview of the research and applications of CLV modeling including an in-depth analysis of the works that have been focused on the F2P field.
In the next section, we will outline a number of foundational concepts that will serve as a basis to the rest of the chapter. In section~\ref{sec:applications}, we will describe different ways in which CLV can be used inside and outside of the game industry.  In section~\ref{sec:models}, we will give an overview of different models used to predict CLV with a more in-depth focus on models employed in the game industry. In section~\ref{sec:software} we will describe a number of software packages that can be used to perform CLV prediction, and finally, in section~\ref{sec:future}, we will outline a number of current and future directions within the field.

\section{Definitions and terminology} \label{sec:foundations}

Pfeifer et al.~\cite{Pfeifer2004}, in their 2004 article, describe a problem present across the different research works on customer lifetime value: incoherent terminology. The reasons behind this problem are probably manifold, e.g. the different research communities; however, in this section we will attempt to synthesize a number of concepts and terms that are common across the different fields involved and are foundational to the study of customer lifetime value prediction in games and beyond.

First of all, it is important to define what is customer lifetime value; Pfeifer et al.~\cite{Pfeifer2004} summarize it as: ``the present value of the future cash flows attributed to the customer relationship''. As the authors point out, this definition makes a couple of important assumptions: first, the value is associated to the \textit{cash flow}, which means it is not limited to the inbound flow of revenue generated by a customer but also to the costs attributed to that customer; however, the costs considered in this definition are only the ones directly attributable to a given customer.
Second, the term \textit{value} expresses a valuation at the current point in time of some future revenues and costs, this implies some form of discount based on the time in which the future cash flows are predicted.

In their definition, Pfeifer et al. do not attempt to encompass all possible definitions of CLV within the literature; instead, they propose one possible explicit definition of CLV that is terminologically correct and coherent. This means that other research works on predicting CLV do not necessarily adhere to the aforementioned definition.
Berger and Nasr~\cite{Berger1998}, for instance, do not consider acquisition cost as part of the lifetime value, considering instead CLV to be the ``maximum profitable acquisition cost''~\cite{Jain2002}.
Sifa et al.~\cite{Sifa2015}, on the other hand, do not take into consideration the present value and use no factor to discount the future predicted revenue.

One further aspect of this definition that is not shared by the different research works is the time frame considered for the calculation. Gupta et al.~\cite{Gupta2006a}, in their review, find that a number of researchers employ an arbitrary time horizon for the future prediction~\cite{Reinartz2000}, while others user an infinite horizon~\cite{Fader2005}. While it might appear that employing a fixed time horizon is just a way to simplify the modeling process, both approaches have their own merits. 
Using an infinite horizon does permits to avoid making any assumption about the maximum length of the customer relationship with the company, therefore; it allows a more accurate prediction of Pfeifer's definition of CLV; however, often for budgeting or other purposes, it is important to know the return on investment within a given period of time (e.g. 1 year), making a fixed horizon CLV prediction extremely useful.

Another consideration relative to the temporal aspect of the customer relationship that has to be taken into account when studying CLV prediction is whether the relationship is exclusive or it allows the customer to use other competitors' services; these two different type of relationship are often labeled as either \textit{lost-for-good} or \textit{always-a-share}~\cite{Dwyer1997}. 
Identifying which type of relationship is being modeled for CLV prediction is extremely important as it affects the temporal length of the relationship and the definition of \textit{churn} (i.e. when a customer stops using the service).

This is especially true in free-to-play-games where there is no visible churn event; in this context, the definition of the user state at any given moment (active or inactive) depends on an ad-hoc formula based one some synthesis of the players' actions~\cite{Runge2014a}.
Many CLV prediction models include or are built on top of a churn prediction model, and both the formula and the choice of model will depend on whether it is possible for the customer to return after a period of inactivity~\cite{Pfeifer2000} or whether his or her relationship with the company is considered finished~\cite{Glady2009}.

Finally, it is important to specify that all of the works cited and described in this chapter do not consider the effects of the competition with other services on the customer lifetime value. As pointed out by Gupta et al.~\cite{Gupta2006a}, this is the case for most current modeling approaches because of the lack of data about the competitors.

\section{Applications of CLV prediction} \label{sec:applications}

While marketing serves as the primary application area of customer lifetime value prediction, a number of other activities, such as customer relationship management or live game operations, are being increasingly driven by data and different key performance indicators (KPIs) such as CLV.
Schmittlein et al.~\cite{Schmittlein1987}, in one of the earliest works on customer lifetime value estimation, motivate their research by stating that: 
\begin{quotation}
"...the issue is important in at least three settings: monitoring the size and growth rate of a firm's ongoing customer base, evaluating a new product's success based on the pattern of trial and repeat purchases, and targeting a subgroup of customers for advertising and promotions."
\end{quotation}
They envision that, by knowing which customer are active and what purchases they make, a manager is able to do more accurate budgeting, a better product evaluation and target the customers more accurately with re-engagement initiatives.
Similarly to Schmittlein et al., a number of other research works have tackled the above problems; Table~\ref{tab:functions} gives and overview of the main activities for which CLV prediction has been employed. 

Each application area includes a number of different activities (e.g. special offers or advertisement) united by a similar usage of CLV prediction. 
In the case of budgeting and finance, CLV predictions is used to predict revenue and plan budgets for the different activities within the company~\cite{Schmittlein1987,Donkers2003}, finding the balance between acquisition and retention investments~\cite{Blattberg1996} or to make a financial estimation of the company~\cite{Gupta2006}.
The second activity, Product Development, is mentioned only by Schmittlein et al.~\cite{Schmittlein1987}; however, as we will describe later in this section, CLV is an important KPI in the evaluation of the game quality in F2P games and it can be very useful to drive the game design process.

\subsection{Customer Acquisition}

Customer Acquisition encompasses all activities aimed at getting new customers to start using the service or buying the products offered. As mentioned by Berger et al.~\cite{Berger1998,Berger2003}, having an estimation of the customers' lifetime value can help deciding more accurately how much to spend on a given promotional campaign and how much margin for profitability there could be in a new market segment; Dwyer~\cite{Dwyer1997} further develops this concepts by identifying in CLV the ceiling of the customer acquisition spend.
Having an early indication of whether the costs of acquisition are matched by the CLVs, can help deciding whether a certain promotional initiative should be continued or stopped~\cite{Lange2014,Sifa2015,Sifa2018}.

\subsection{Customer Retention}

Customer retention and segmentation initiatives are often intertwined and not completely distinguishable; in this article, we describe customer retention/loyalty activities as any activity explicitly aimed at prolonging the lifetime of a customer, while, with customer targeting/segmentation, we identify all activities aimed at identifying different homogeneous groups within the customer base that can be targeted with a custom user experience. Such a customer experience could be aimed at prolonging the customer lifetime; however, for the purpose of this categorization, we define this as a secondary objective.

\begin{table}
\centering
\begin{tabular}{|l|l|}
\hline 
\textbf{Application area} & \textbf{Article} \\ 
\hline 
Budgeting and Finance & \cite{Schmittlein1987,Blattberg1996, Donkers2003,Gupta2006} \\ 
\hline 
Product Development &  \cite{Schmittlein1987}\\ 
\hline 
Customer Acquisition & \cite{Dwyer1997, Berger1998,Berger2003,Lange2014,Sifa2015, Sifa2018} \\ 
\hline 
Customer Retention/Loyalty & \cite{Berger1998,Mulhern1999,Reinartz2000,Lachowetz2001,Rosset2002,Rosset2003} \\
& \cite{Malthouse2005,Shen2009,Cheng2012,Runge2014a,Sifa2018} \\ 
\hline 
Customer Targeting/Segmentation & \cite{Mulhern1999,Reinartz2000,Verhoef2001,Venkatesan2004,Hwang2004, Haenlein2007, Shen2009, Kumar2010, Khajvand2011} \\ 
\hline 
Other &  \cite{Aeron2008}\\ 
\hline 
\end{tabular} 
\label{tab:functions}
\caption{Applications of CLV prediction in different business activities}
\end{table}

Latchowetz et al., in their study on the impact of season ticket holders on the NBA franchises revenue~\cite{Lachowetz2001}, argue that a customer should not be evaluated solely on the revenue that she or he generates within a season. At the time of their estimation, an average season ticket holder had a lifetime value of more than eighty thousand US dollars; therefore, the authors advice the entertainment industry to acknowledge the importance of using customer lifetime value and developing long term retention strategies for their customers.

Berger and Nasr~\cite{Berger1998} discuss how knowing the customer lifetime value can help determining the effects of adopting different marketing strategies for retention and acquisition and how to balance the spending between the two activities. The authors point out also that any acquisition strategy might impact also the customers' retention; therefore, it is hard to see the two activities independently and both their budgets must by aligned to the predicted customer lifetime value. Mulhern extends further Berger and Nasr's idea about the strategic use of CLV prediction by seeing it as a primary indicator to driving resources allocation in the marketing mix~\cite{Mulhern1999}.

Rosset at al.\cite{Rosset2002,Rosset2003} extend the idea of using CLV prediction to drive retention activities and expand it by investigating how to estimate the impact of retention efforts to the CLV. They demonstrate an application of their approach in the context of a retention campaign for a group of potential churners. They show how, by having a model able to predict CLV in different configurations of the service, it is possible to compare the current estimated lifetime value of a group of potential churners with different projected lifetime values given a number of different incentives to improve the cohort's retention; thus, taking a more informed decision on the correct initiative to be taken.

The predictions used for these decisions, as any other prediction of future events, have some degree for uncertainty in the form of prediction error or bias; taking this uncertainty into account can be also important in decision making. Malhouse and Blattberg~\cite{Malthouse2005} study the impact of uncertainty in the management of the customer relationship helping to understand not only whether engaging into a retention activity can be profitable but also how often it could be.

\subsection{Customer Segmentation}

Most of the research works presented in this section make the assumption that a higher retention (i.e. a longer lifetime for the customers) will have a positive impact on the company's revenue in the form of an increased CLV; however, while this is generally true, Reinartz and Kumar~\cite{Reinartz2000} found that this assumption does not always howl and different customer segments have different spending patterns, which means that, in a number of cases, prolonging the customer's lifetime would no yield any more revenue. 
For this reason, they suggest that, the company should not necessarily pursuit long-term customer relationship, but it should customize the relationship based on prediction on future customer lifetime value and retention.

Segmenting the target audience and personalizing marketing activities is an established practice~\cite{Dickson1987}; however, the predominant guiding factors used to define the segments and the most appropriate actions have, for long time, been based on demographics, questionnaires and interviews~\cite{Malhotra2008}. The advent of large scale data collection and analytics, in many industries, has dramatically changed many of the common practices in marketing research, allowing practitioners to access more detailed information from wider audiences at lowers costs.

Mulhern~\cite{Mulhern1999}, for example, proposes customer lifetime value as a possible segmentation criterion beside usage volume and brand loyalty: he suggests that one simple segmentation could divide the users in customers worth keeping, as their CLV is profitable, and customers that are not worth the retention cost. Verhoef and Donkers~\cite{Verhoef2001} propose a similar segmentation to design personalized insurance offers. 

Venkatesan and Kumar~\cite{Venkatesan2004} pose the question whether using predicted CLV for segmentation con yield better results than more established KPIs such as previous-period customer revenue or past customer value. To compare the effectiveness of the different metrics, they rank the customers based on each metric and segment the customers based on the raking; the results show the segmentation based on CLV to be superior to the other segmentation approaches.

Similar strategies to the ones mentioned so far have been applied to a number of different industries such as telecommunications~\cite{Hwang2004}, banking~\cite{Haenlein2007}, retailing~\cite{Shen2009, Kumar2010} or beauty~\cite{Khajvand2011}.
In the last few years, a similar trend has emerged in the gaming industry and, especially in the mobile game and free-to-play industry, all the aforementioned marketing practices are established and widely undertaken~\cite{Shankar2009}.

\subsection{Free-To-Play Games}

\begin{figure}
\centering
\includegraphics[width=3.5cm]{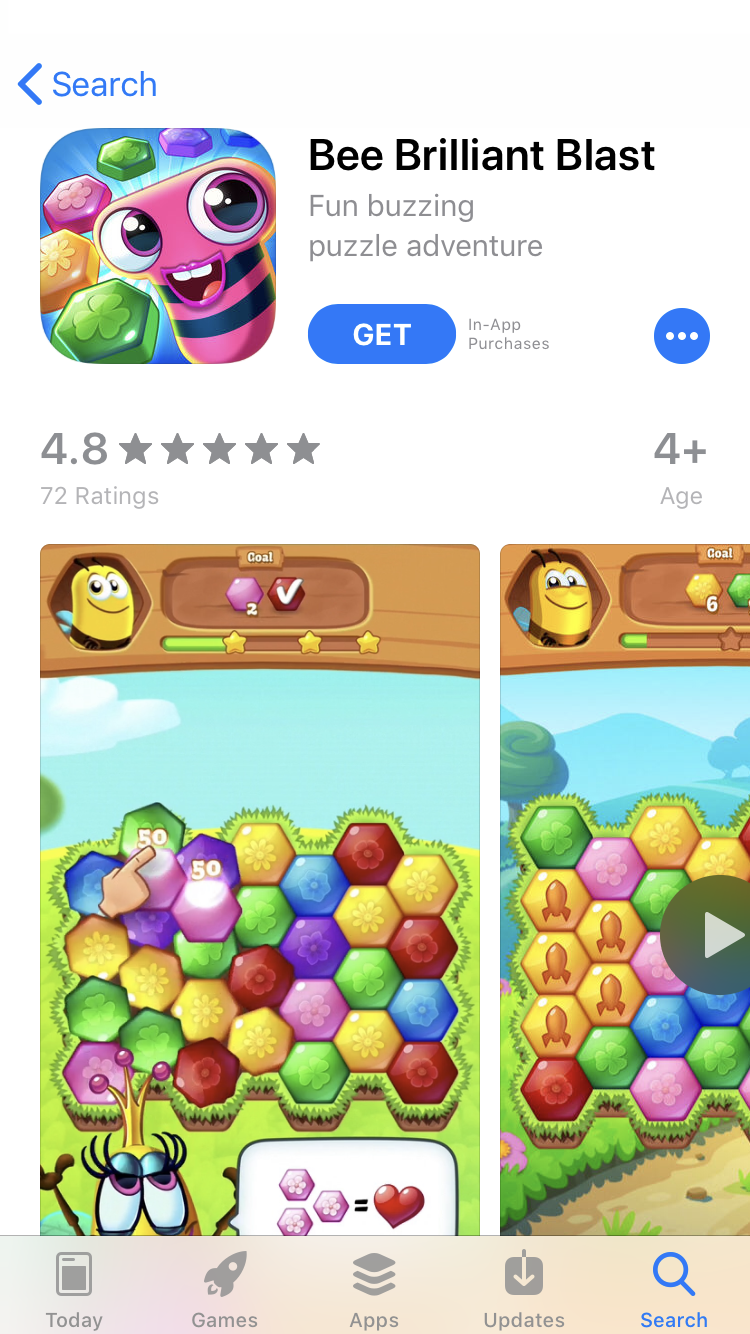} \ \
\includegraphics[width=3.5cm]{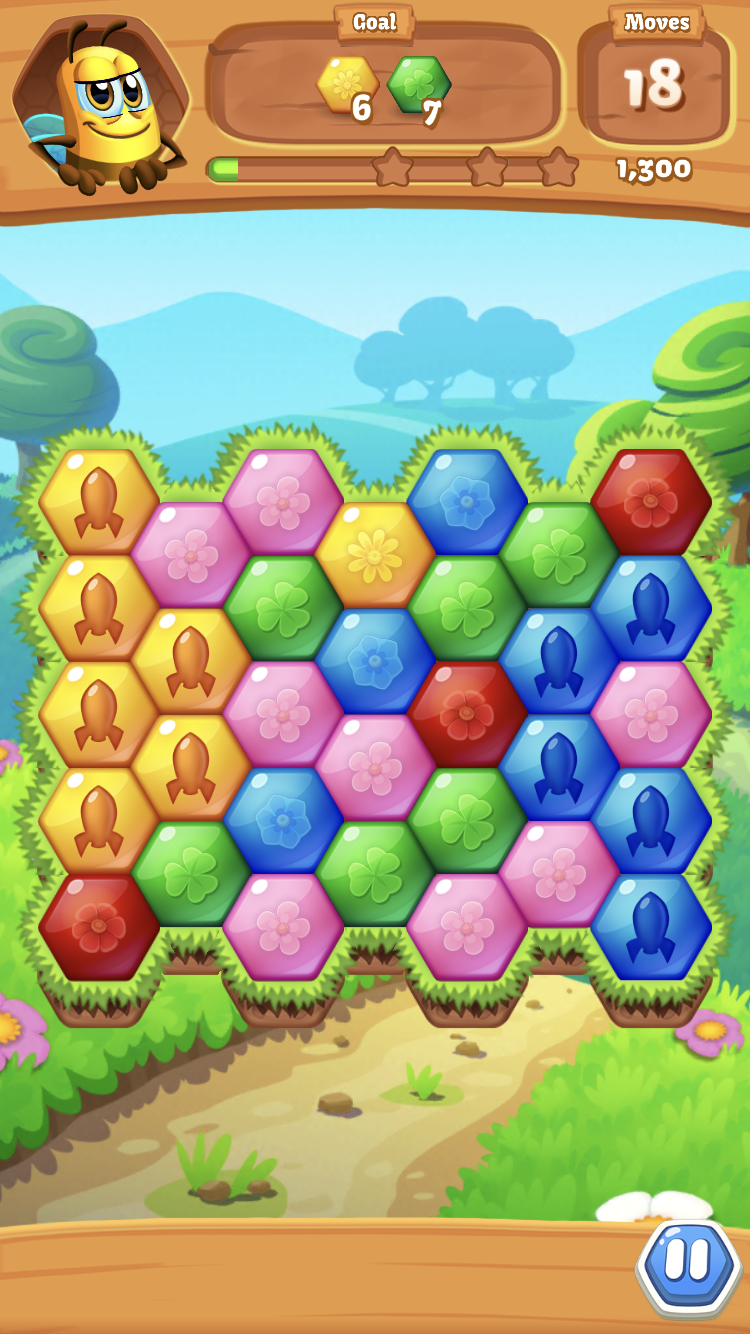} \ \
\includegraphics[width=3.5cm]{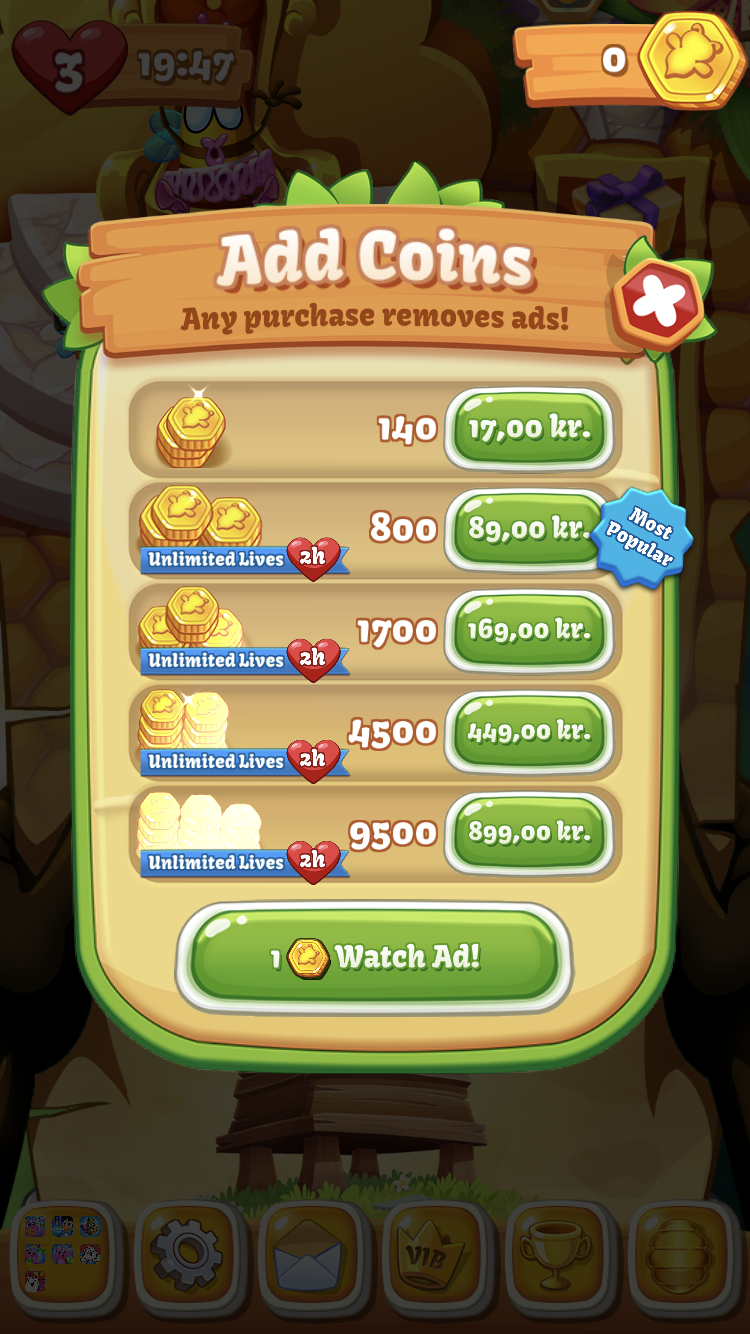} \ \
\caption{Three screen grabs from the iOS version of Bee Brilliant Blast by Tactile Games depicting, from left to right, the phone's store interface, a typical puzzle level and the in-game store interface, in which the customer can purchase different amounts of in-game currency.}
\label{fig:bbb}
\end{figure}

Free-To-Play video games are games than can be played without an upfront payment and potentially completely free of charge; players can, within the game, purchase some virtual goods and pay to enable some features. While this type of games was originally mostly distributed through on-line social media platforms, the business model is currently vastly diffused on many different platforms, such as mobile phones, home consoles and personal computers~\cite{Alha2014}.

Figure \ref{fig:bbb} shows an example of a mobile free-to-play game; as it is possible to see on the first screen grab to the left, the game is freely available on the phone's store, however, the store specifies that the game contains the possibility to perform in-app purchases. The last image to the right shows the in-game store that allows the customer to purchase different packs of in-game currency, which can be used throughout the game to unlock different features. Using an intermediary currency, while a very common practice, is not defining characteristic of free-to-play games, and some games allow customer to enable directly in-game features directly using a real currency.

The second main source of revenue in free-to-play games comes from advertisement: a large portion of free-to-play game display to the players ads in different formats (e.g. videos or interstitials). These ads are provided by external services that act as mediators between the advertiser and the publisher. The game (i.e. the publisher) received ads from the ad provider through an API and shows them to the player, the ad provider will in return pay the game developer depending on the number of ads being displayed, the number of times the ads are clicked or the number of times the application being advertised is being installed.

Both aforementioned revenue streams -- in-app purchase and advertisement -- have a non-contractual nature, as the player can freely choose if and when to make a purchase or click on and advertisement campaign; furthermore, a conversion rate (i.e. the percentage of players making an in-app purchase) well below 10\% is considered normal in the industry~\cite{Nieborg2015}, making it especially challenging to predict the customers' lifetime value. At the same activities such as customer acquisition and retention rely on accurate CLV predictions for accurate targeting and budgeting.

As of November 2018, there are around 816000 games available only on the Apple App Store\footnote{https://www.pocketgamer.biz/metrics/app-store/categories/}; therefore, targeting the right customers and accurately measure their potential impact in the company's revenue is particularly important in customer acquisition in the free-to-play games market, given the extremely high number of games competing for the same users.
A significant part of the customer acquisition efforts in the industry is executed through performance marketing campaigns: in these campaigns, he advertiser competes with other advertisers to show ads the potential new customers, often times through some form of auctioning system. 
Customer lifetime value predictions can be used as a benchmark to evaluate the profit margin of these campaigns as both the potential revenue coming from the new customers and their cost of acquisition vary greatly depending on the segmentation chosen for the campaign, the market context and the quality of the advertisement creative content~\cite{Sifa2018}.
Even beyond performance marketing, an accurate CLV prediction is very important to be able to budget expensive and higher risk marketing efforts such as television advertisement.

Furthermore, the F2P games market is characterized by very low retention number -- e.g.  the best android games have a day 30 retention rate below 5\%\footnote{https://www.forbes.com/sites/johnkoetsier/2017/10/20/the-very-best-android-game-has-just-4-5-user-retention-after-30-days/} -- making any effort to improve such low number a priority for most companies. Within this context, CLV prediction can be used, as previously described, to target different segments of users with special offers; However, it is also possible to personalize the game experience at large by adapting, for instance, the flow of the game progression or by providing custom events. Harrison and Roberts show an example of such adaptation aimed to improving retention in a digital version of Scrabble~\cite{Harrison2013}.

Given the importance of CLV prediction in the free-to-play and other industries, a number of algorithms have been developed over time , in the next two section we will describe the most significant ones and analyses a number of software packages that can be used for CL prediction.

\section{Models} \label{sec:models}

Berger and Nasr are the first researchers that attempted a categorization of CLV prediction models~\cite{Berger1998}. In this early review of the field, all methods mentioned are aimed at calculating an average customer lifetime value for the entirety of the company's customer base or a given cohort.
In a latter review~\cite{Kumar2004}, Kumar et al. group these approaches under the label "average LTV approach", while Jain and Singh~\cite{Jain2002} use the term "basic structural model".
In this chapter we adopt the definition used by Kumar et al. and we include a number of other practices commonly used in the free-to-play industry that approach LTV prediction in a similar fashion.

The methods for CLV prediction discussed in this article are dividend in two further categories: customer history models and machine learning models. The first category contains all methods that attempt to predict either a single customer's or a cohort's lifetime value based on some mathematical projection if the past behavior of the given customer or cohort. The second category includes all algorithms which make use of some learning algorithm to build a computational model of the customer behavior, which is then used to make predictions. 

\subsection{Average Models}

The aim of the models discussed in this section is to give an estimation of the future discounted cash flow for a group of customers or for the whole user base. The common aspect between these models is that they make a number of assumptions: they assume a constant (or otherwise known) margin of profit over time, they do not consider the stochastic nature of the purchase behavior of the customers and they assume that the customer behavior is uniform across the estimated cohort.

The most basic model proposed by Berger and Nasr~\cite{Berger1998} assumes that the customer retention rate and the costs of retention are constant over time and both costs and revenues happen periodically at a constant rate, within these conditions the formula for CLV is as follows:

\begin{equation}
CLV = GC * \sum_{i=0}^n \frac{r^i}{(1+d)^i} - M * \sum_{i=0}^n \frac{r^{i-1}}{(1+d)^{i-0.5}}
\label{eq:basicltv}
\end{equation}

Where \textit{CG} is the customer's expected gross contribution margin per period, \textit{M} is the promotion costs per customer per period, \textit{n} is the prediction horizon expressed in number of periods, \textit{r} is the retention between one period and the next, and \textit{d} is the discount rate. 

Further iterations of this model presented by Berger and Nasr~\cite{Berger1998}, by Jain and Singh~\cite{Jain2002} and by Kumar et al.~\cite{Kumar2004}, allow more sophisticated ways to express non constant retention rates and margins of profit and to include acquisition costs in the calculation.

In their article about the relationship between the length of service and CLV, Rosset et al.~\cite{Rosset2003} formulate an extended average model that employs a Kaplan-Meier~\cite{Kaplan1958} estimator to calculate the duration of the relationship between the company and the average customer . 

A similar approach is currently among the basic models used in the free-to-play industry for CLV prediction~\cite{Seufert2013}: after selecting a specific function to express the retention curve of a given cohort, the function is fit on the training data and the resulting customer lifetime value is calculated as follows:
\begin{equation}
CLV = \sum_{i=0}^n ARPDAU * ret(i)
\end{equation}

Where $ARPDAU$ stands for average revenue per daily active user and $ret(i)$ is the value of the chosen retention function at the $i^{th}$ day. This method for CLV prediction, while quite simplistic and not necessarily extremely accurate, has the advantage of being easily readable and based on established business KPIs such as APRDAU and retention.

Another example of simplistic but widely adopted model is described by Runge~\cite{Runge2014}: instead of basing the projected revenue on retention and APRDAU, the method fits a function to approximate the average monetization curve, which is then used to project the customer lifetime value for a player based the amount of money the she spent at the present day.
If we consider $rev(n)$ as the revenue produced by a customer up the $n^{th}$ day of the customer relationship and $mon(n)$ as the average fraction of the revenue that is produced by day $n$ the formula for CLV prediction is as follows:

\begin{equation}
CLV = rev(n)/mon(n)
\end{equation}

The major advantage of this model is its simplicity and, similarly to the retention-based model, its readability. However, with such a model it is not possible to predict the CLV for customers that have not yet produced any revenue -- i.e. the projection would always result 0; moreover, in a context such as free-to-play games, in which the amount of paying user is very small, the variably of the behavior between different paying user has a large impact on the prediction, and such variability is completely disregarded by the model.

\subsection{Customer History Models}

The aforementioned methods based on retention and monetization curves are partly based on historical customer data as, in both methods, a function representing the average retention or monetization behaviors is fit on some past customer data and then used for prediction. These methods however do not allow to use any particular user's track record to personalize the prediction, hence the average nature of the methods.

Based on the customer categorization by Jackson~\cite{Jackson1985} that divides the customers in two types: \textit{always-a-share} and \textit{lost-for-good}, Dwyer~\cite{Dwyer1997} argues that an average retention model is not sufficient to model CLV for customers belonging to the \textit{always-a-share} category. For this type of customers, he proposed a migration model based on the customers' recency of purchase.

\textit{Always-a-share} customers differ from \textit{lost-for-good} customers in that the latter have a long-term commitment with the company and switching to a competitor is costly, while the first ones can have simultaneous relationships with multiple companies competing for the same product/service. Typical examples of \textit{lost-for-good} customers are bank customers or tenants in a rental property. Whether free-to-play game players can be considered belonging to the first or second category is an open question as there is no monetary barrier that stops a player to switch to another game; however, to a certain extent, the longer the time a player invests in a game, the less likely is that the player will be stopping playing that game.

In his customer migration model, Dwyer uses purchase recency to estimate the customer's probability of performing a purchase in the next period and the potential value of such a purchase. In the training phase, based on past customers' data, the method builds a model of purchase probabilities and values that depend on the length of the period of inactivity. These recency cells are then used as a lookup table to predict, depending on each customer's past purchase behavior data, her purchases in the next time unit.
Dwyer's model allows to leverage the customer's past behavior for a personalized CLV prediction and considers the probabilistic nature of the customers purchase behavior. 

\subsubsection{Recency, Frequency and Monetary Value}

The main limitation of Dwyer's methods is that the probability of a purchase in a given period is based only in the recency of the last purchase, which is not necessarily true for all businesses.  Especially in free-to-play game, different types of purchase will affect differently the future purchases, for example, a player purchasing a large package of virtual currency recently might be less likely to make purchase as the amount of resources acquired will allow her to play longer without a need for any help within the game.

The concept of recency used by Dwyer is an established metric in direct marketing and alongside purchase frequency and average purchase monetary value -- commonly known together as RFM -- have been long used in the field to predict customer behavior~\cite{Gupta2006a}.
Hughes describes a method for customer quality estimation based on these three variables: the current customer based of the company is sorted along these three variables each categories into 5 quantiles, creating therefore $5 * 5 * 5$ groups, these groups are then used to score the different customers and target them with specific offers~\cite{Hughes2000}.

Inspired by Hughes's work, Shih and Liu propose a method based on RFM and CLV clustering to rank the customer according to their profitability~\cite{Shih2003}. As a first step the method relies on a group of expert evaluation to identify the relative importance of the recency, frequency and monetary variables using analytical hierarchical processing. The customers are than clustered based on the RFM space and the resulting clusters are ranked through a simple weighted sum of the three normalized variables. 

The aforementioned methods are not designed to produce a numerical predictor of CLV but only to score the customers by their potential profitability.  Furthermore, these methods are limited in predicting the customer behavior only for one future time period~\cite{Fader2005a}. Finally, these methods disregard the fact that customers' pas behavior is, often times, the result of past company activities.

Fader et al. describe a model for CLV prediction, using RFM as input variables based on the e Pareto/NBD framework~\cite{Schmittlein1987}, which overcomes some of the aforementioned limitations~\cite{Fader2005a}.

\subsubsection{Pareto/NBD}\label{sec:btyd}

The Pareto/NBD model~\cite{Schmittlein1987} aims at predicting individual customer's purchase behavior based on their past purchase patterns. More specifically, for each customer, based on the recency ($t$) and frequency ($X$) of the purchases and the length of the relationship between the customer and the company ($T$), the model estimates the expected future number of purchases.
The model is based on five assumptions:
\begin{itemize}
\item while active, a customer makes purchases following a Poisson process with rate $\lambda$;
\item the purchase rate $\lambda$ differs between customers and is distributed according to a gamma distribution across the customer base;
\item the duration of the active period of the customer is exponentially distributed with a death rate $\mu$;
\item the death rate $\mu$ differs between customers and is distributed according to a gamma distribution across the customer base;
\item the death rate and the purchase rates are independent.
\end{itemize}

Based in these assumptions, the authors find that the customers ``deaths'' for a sample follow a Pareto distribution of the second kind~\cite{Johnson1970}, while the number of purchases made by an active customer follow a negative binomial distribution~\cite{Ehrenberg1972} -- hence the name Pareto/NBD. The two distributions are controlled by four parameters ($r$ and $\alpha$ for NBD and $s$ and $\beta$ fro Pareto); 

Schmittlein suggests two approaches to estimate the parameters based on the customers' past behavior: maximum likelihood and fitting observed moments; in case a model in needed without prior data, Schmittlein discusses also the possibility to handpick the parameters based on management judgments.
Once the parameters have been identified, each customer's future number of purchases is predicted using the two aforementioned distributions and the customer's $X$, $t$ and $T$.

One of the main issues with the Pareto/NBD is it's computational complexity~\cite{Fader2005} as the estimation requires multiple evaluations of the Gauss Hypergeometric Function; this problem is mitigated by the modified BG/NBD model by Fader et al.~\cite{Fader2005}, which models the customer activity using a beta-geometric model that is easier to implement efficiently.

Both aforementioned models are able to predict the future number of purchases for a given customer and they can, for instance, be used to count expected active users ate a give point in time, however these methods do not model in any way the value of each purchase, therefore, they can't directly predict the customer lifetime value. 

Reinartz and Kumar, in their studies on profitable lifetime duration~\cite{Reinartz2000, Reinartz2003}, employ Schmittlein's model to be able to calculate the number of time periods in which a customer will perform a purchase. 
The authors transform the continuous probability that a customer is active into a dichotomous measure of whether the customer is active or inactive it a given point in time. Given a probability threshold, it is possible to identify a customer's future date of churn, which, combined with the first date of the customer relationship, gives an expected customer relationship duration.
This duration, expressed as number of periods $n$, is used to calculate that customer's lifetime value using the formula defined by Berger and Nasr~\cite{Berger1998} and described in Equation \ref{eq:basicltv}.

Schmittlein and Peterson\cite{Schmittlein1994} propose an extension of the original Pareto/NBD in which the future monetary value of each transaction is samples from a normal distributed around the mean monetary value of a cohort. However, Fader et al.~\cite{Fader2005a} observe that, in the data they analyzed, there are large differences between mean, mode and median, indicating that the distribution of the monetary values is highly skewed. Therefore, the authors propose to model the average transaction value using an adapted version of the gamma-gamma model proposed by Colombo and Jiang~\cite{Colombo1999}.

One of the main strengths and, at the same, the main limitations of the Pareto/NBD or BG/NBD models is that they rely on a small number of purely transactional data. This is a strength in that it makes the model general enough to be easily applied in different context; however, the resulting model will always risk to be sub-optimal as it disregards potentially relevant information. Sing et al. address this limitation and propose an estimation framework that can flexibly incorporate multiple statistical distributions and consider a number of covariates such as age or gender ~\cite{Singh2007}.

Glady et al.~\cite{Glady2009} address another important limitation of the aforementioned models: they all assume independence between the frequency of transactions and the profit per transaction. The authors of the study demonstrate that such assumption does not hold in multiple real-world data sets and propose a modified Pareto/Dependent model that performs better than Pareto/NBD in such circumstances.

\subsection{Markov Chain Models}\label{sec:mcm}

An alternative approach to customer lifetime value prediction is proposed for the first time by Pfeifer and Carraway~\cite{Pfeifer2000}: the authors suggest to model the customer relationship as a Markov Chain Model (MCM), in which the different states of the model represent different conditions of the relationship between the customer and the company in terms of transactions and customer activity, and the transition probabilities between states represent the probability of a customer to move from one condition to the other one -- e.g. for a customer to make a purchase or to churn. 

\begin{figure}
\centering
%   \digraph[width=12cm]{abc}{
%      rankdir=LR;
%      purchase -> r1 -> r2 -> r3 -> r4 -> churned;
%      r1 -> purchase;
%      r2 -> purchase;
%      r3 -> purchase;
%      r4 -> purchase;
%   }
   \includegraphics[width=12cm]{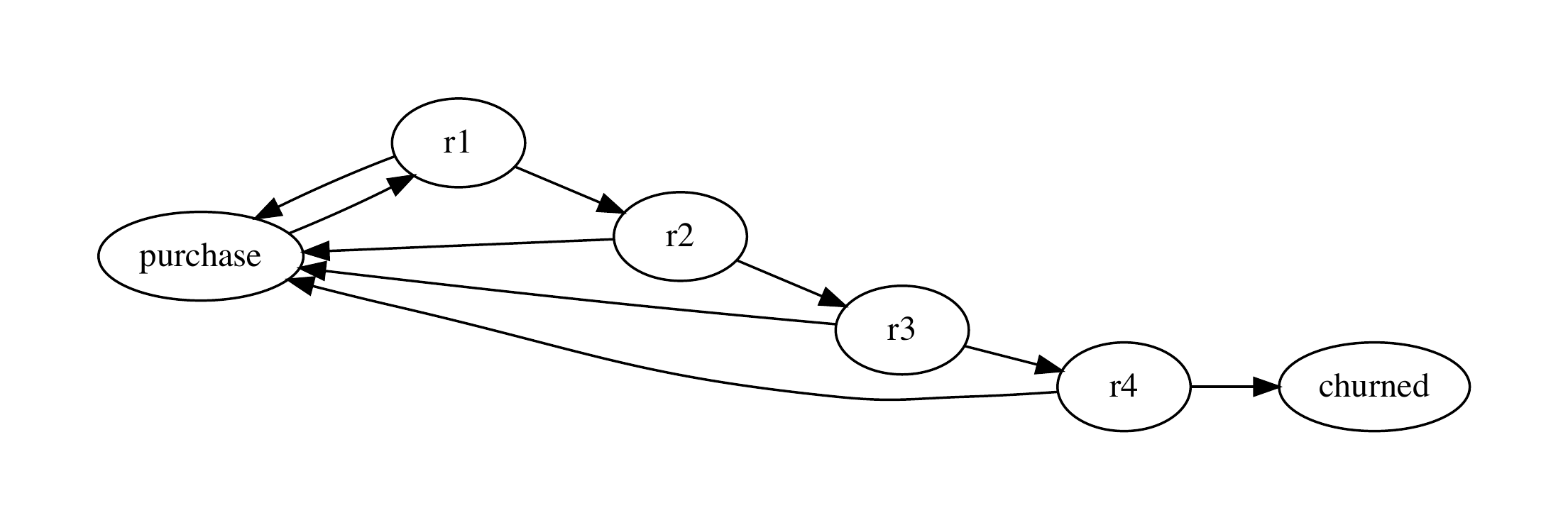}
   \caption{Simple customer relationship Markov Chain Model. States r1 to r4 represent different value of recency from 1 to 4.}
   \label{fig:markov}
\end{figure}

Figure \ref{fig:markov} shows an example of a simple Markov Chain Model representing the possible states of the relationship between a customer and the company and the valid transitions between the states. 
In the depicted case, the states are based on the recency of the last purchase (e.g. r1 corresponds to recency equal to 1) and there are only 5 valid stages before churn. However, as in the examples reported by Pfeifer and Carraway, in a real-world scenario, the states are based on a multitude of factors and there are many more valid transitions each with their own probability. 

While Pfeifer and Carraway show how it is possible to calculate CLV using a MCM representation of the customer relationship with a given set of states and transaction probabilities, Etzion et al. focus on the process of learning the states and the transition from past customer data~\cite{Etzion2004}. The process requires first to identify the variables determining the states of the model, to define their ranges and discretize them. 

At this point the transition probabilities between the states can be calculated through the following three steps:
\begin{enumerate}
\item initialize a transition matrix -- i.e. a square matrix that contains the probability of transitioning between one state and another one -- with zeros in all cells;
\item for each customer performing a transition between a state $i$ and a state $j$, the matrix cells identified by $ij$ is incremented.
\item at the end of the processing, each line of the matrix is normalized between 0 and 1 using a min-max normalization.
\end{enumerate}

The resulting matrix can be used, as suggested by Pfeifer and Carraway~\cite{Pfeifer2000} to calculate CLV for a customer in any given state. 

Ching et al.~\cite{Ching2004} employ the same approach to build multiple transition matrices that describe the customer behaviors in different market conditions -- e.g. with or without a promotion -- finding that the transition matrices differ and so does the CLV and the customer retention. The authors further showcase how to use stochastic dynamic programming to estimate the optimal promotion strategy that optimizes CLV given the previously found transition matrices and a set of constraint on the promotions budget.

\subsection{Supervised Learning Models}\label{sec:supervised}

The general principle at the core of all supervised learning algorithms is to learn a function between some example input and output data -- e.g. past customer behavior and their recorded lifetime value. This example data is usually called training data, and the function resulting from this training phase -- the model -- is used to predict the target variable -- e.g. CLV -- on newly collected data.
This procedure is not dissimilar to some of the MCM based approaches previously described~\cite{Etzion2004}; however, the models we present in this section, differ from MCM models in that they trade explanatory power for the possibility to capture more complex behaviors using some form of computational/black-box model. 

This means that the resulting models are less useful to inform the business about specific customers' preferences, but they can potentially be more accurate and be deployed, for instance, for marketing automation. 
It is important to note that this is not a hard distinction and many models described in this and the previous sections include a mix of explanatory and black-box models. 

An example of a model incorporating different learning algorithms is the one proposed by Haenlein et al.~\cite{Haenlein2007}. The authors describe an approach to CLV prediction that uses a combination of CART analysis~\cite{Breiman2017} and MCM: first, the customers of a retail bank are divided into age groups and, for each age group, a regression tree model is built to predict the profit for a customer within that group; second, the transition probabilities between the groups are modeled as a Markov Chain so that the model is able to follow the transitions of the customers throughout their lifetime. The resulting lifetime value is calculated as a discounted sum of each of the CLV predicted in the possible customer states weighted by the transition probabilities. 

Cheng at al.~\cite{Cheng2012} approach CLV prediction in a similar way but they extend the model by building a different Markov Chain for each period the customer relationship; each chain includes a number of states depending of the activity level of the customers in that specific period. The total CLV for a customer is similarly calculated by summing the predicted CLVs over the different states and periods weighted by the transition probabilities and the survival probability of each period. The models used to predict the profit contribution of a customer and a given state is modeled using an artificial neural network, while the survival probability in each period is calculated using a logistic regression model.

Within the context of free-to-play games, supervised learning methods have been employed in a number of studies. Compared to the previously mentioned approaches, supervised learning offers two key advantages that make them particularly interesting within this industry: the ability to leverage a wide variety of data about the customer behavior and the ability to make predictions about CLV for players that are not yet customers -- i.e. that have not yet made a single purchase.

The first aspect is important as in the freemium games industry, customers are also players and the data describing the gaming behavior of players/customers is much richer and at a much higher frequency compared to their purchase behavior data. 
Furthermore, there are a number of studies that show how past in-game behavior is a predictor of future in-game behavior~\cite{Mahlmann2010cig}, of player preferences~\cite{Burelli2015umuai} and of the probability of making a purchase within the game~\cite{Hanner2015a}.

The second aspect is partially related to the first one: since the predictions can be based also on non-purchase related data, it is potentially possible to perform predictions also for players that have not yet made a purchase and that might never be making one. 
As described in Section \ref{sec:applications}, most of free-to-play players do not make any in-game purchase and generate little to no revenue; it is therefore important for a model that predicts CLV in such context to be able to discern which customers will be making a purchase before the first one is made.

In one of the first studies on the application of supervised learning for CLV prediction in freemium games, Lange et al. present a comparison  of the performance of a series of linear and non-linear models in the prediction of CLV~\cite{Lange2014}. The dataset used for the study contains in-game events from the first 7 days of gameplay of about 38 million players; the input variables include data about player spending, gameplay, game progression, social interactions, success metrics and game settings preferences. The target variable is the amount of revenue generated by each player after 180 days.

The results of the study show that an ensemble of two artificial neural networks using the absolute-differences technique over-performed the other methods in terms of mean square error in a 10-fold cross-validated comparison.

A similar study is presented by Sifa et al.~\cite{Sifa2015} in their article on purchase behavior prediction in mobile free-to-play games. Similarly to Lange et al., the authors present a comparative study of multiple supervised learning algorithms; however, the objective of the prediction is slightly different as the authors argue that CLV prediction is a combination of three prediction tasks: predicting whether player will ever make a purchase, predict the number of purchases and predict the value of each purchase.
The results of the study show that, for the classification task, the models having the best performance are tree-based models -- e.g. random forest and decision tree classifiers -- and that it is extremely valuable to re-sample the dataset in to avoid as the players performing purchases are extremely under represented. For this purpose, the authors employ the SMOTE-NC~\cite{Chawla2002} method to over-sample the ``converted'' users data points.

In a latter article, Sifa et al. present a study that attempts to predict CLV in one single step~\cite{Sifa2018}, more in line with the approach presented by Lange et al.~\cite{Lange2014}. The study evaluates the performance of a deep neural network in predicting day 360 cumulative undiscounted revenue per player, and it compares the results against a number of algorithms such as random forest and linear regression. 
The results of the comparison show that the deep neural network out-performs the other regression algorithms in terms or normalized root mean square error.

Furthermore, the authors demonstrate how to adapt SMOTE-NC to re-sample the dataset and reduce the imbalance between customers that made a purchase and customers that did not. This process requires a number of adjustments of the standard implementation of SMOTE-NC, as the original algorithms is design for classification problems, while a direct prediction of CLV is a regression problem.

\section{Software Packages} \label{sec:software}

In Section \ref{sec:models}, we gave an overview of the current state-of-the art in CLV prediction mostly from an academic perspective, listing and explaining a number of scientific articles describing the main approaches to solve the prediction problem. 
However, most of the algorithms presented so far have been translated, to different degrees, in software packages that can be applied to easily perform CLV predictions on new datasets.

In this article, we identify three main packages that can be either used directly to predict CLV or that can be used to construct a model for CLV prediction. The software packages are Lifetimes~\cite{Davidson-Pilon2018} and BTYD~\cite{Dziurzynski2014} that implement some of the algorithms presented in Section \ref{sec:btyd}, and scikit-learn~\cite{Mueller2018} a popular machine learning package for the Python programming language that can be used to implement the models presented in Section \ref{sec:supervised}. Alternative packages can be used to develop supervised learning models -- e.g. Keras~\cite{Chollet2018} or XGBoost~\cite{Cho2018} -- however, most of them are compatible with scikit-learn, therefore we chose this package as the representative in this article.

To build and operate with Markov Chain Models, there are a large number of different alternatives for both the Python and the R language -- e.g. the DTMC pack~\cite{Spedicato2017} and PyMC3~\cite{Salvatier2016} -- however, these packages are designed to offer functionalities that go far beyond the code needed to implement the models presented in Section \ref{sec:mcm}, which can be easily implemented using standard algebra functionalities. For this reason, we chose not cover any MCM package in this Section.

\subsection{Lifetimes/BTYD}
The Lifetimes~\cite{Davidson-Pilon2018} and BTYD~\cite{Dziurzynski2014} packages are both software implementations of the Pareto/NBD, BG/NBD and Gamma-Gamma models; they allow to learn the parameters of the various distributions from past purchases data and to predict future purchases numbers and the purchases values for new customers. In this section we will give sample of how to train and produce predictions with Lifetimes; however, the procedure is very similar for the BTYD software package.

The main components of the Lifetimes package are the classes \textit{BetaGeoFitter} and \textit{GammaGammaFitter}.
The first class (\textit{BetaGeoFitter}) contains all the logic necessary to fit a BG/NBD model from transactional data in the format of purchase recency and frequency, and current age of the customer in terms of customer relationship duration. It is also possible to alternatively fit a Pareto/NBD model using the \textit{ParetoNBDFitter} class instead.

Given a dataset with correctly formatted inputs, the code to fit the \textit{BetaGeoFitter} looks as follows:
\begin{lstlisting}[language=Python, backgroundcolor=\color{lightgray}, frame=line]
from lifetimes import BetaGeoFitter
bgf = BetaGeoFitter(penalizer_coef=0.0)
bgf.fit(data['frequency'], data['recency'], data['T'])
\end{lstlisting}

The package provides a function named \textit{summary_data_from_transaction_data} that allows to produce correctly formatted data for the algorithm from a list of transactions.

The second class (\textit{GammaGammaFitter}) provides all the methods necessary to fit a Gamma-Gamma model and can be used, in combination with the \textit{BetaGeoFitter}, to predict the customer lifetime value. 
The Gamma-Gamma model is also trained on transactional data containing the frequency and the total monetary value of the purchases made by each customer.

The code to fit the Gamma-Gamma model looks as follows:
\begin{lstlisting}[language=Python, backgroundcolor=\color{lightgray}, frame=line]
from lifetimes import GammaGammaFitter
returning_customers_summary = data[data['frequency']>0]
ggf = GammaGammaFitter(penalizer_coef = 0)
ggf.fit(returning_customers_summary['frequency'],
        returning_customers_summary['monetary_value'])
\end{lstlisting}

And the code to perform CLV predictions on new data using the previously fit models looks like this:
\begin{lstlisting}[language=Python, backgroundcolor=\color{lightgray}, frame=line]
ggf.customer_lifetime_value(
    bgf,
    new_data['frequency'],
    new_data['recency'],
    new_data['T'],
    new_data['monetary_value'],
    time=180,
    discount_rate=0.01
\end{lstlisting}

The time and discount rate depend on the parameters of the specific prediction to be performed.

Further documentation and tutorials about the two packages can be found on their project web pages\footnote{https://github.com/CamDavidsonPilon/lifetimes}\footnote{https://cran.r-project.org/web/packages/BTYD/index.html}, and both packages are currently available for download and install respectively on the Python Package Index\footnote{https://pypi.org} and The Comprehensive R Archive Network\footnote{https://cran.r-project.org}.

\subsection{scikit-learn}
Scikit-learn~\cite{Mueller2018} is a library for the Python programming language that provides functionalists to perform supervised and unsupervised learning tasks. The supervised learning part of the library implements a wide variety of algorithms ranging from linear models to tree-based methods for both classification and regression. In this section, we will show how to use scikit-learn to build a regression models based on the Random Forest algorithm similar to the one presented in ~\cite{Lange2014} and~\cite{Sifa2018}.

Random Forest~\cite{Breiman2001} is an ensemble algorithms that combines a number of tree predictors trained on different portions of the data, the output of the model is generally either the mode or the mean of the outputs of the different tree predictors depending on whether the task is a classification or a regression task. Scikit-learn implements this class of algorithms though the \textit{RandomForestRegressor} and \textit{RandomForestClassifier} classes; for the purpose of this article we will show how to use the \textit{RandomForestRegressor} to predict CLV through regression.

Given a dataset similar to the one used in~\cite{Sifa2018} with one line per player, the code to fit the \textit{RandomForestRegressor} looks as follows:
\begin{lstlisting}[language=Python, backgroundcolor=\color{lightgray}, frame=line]
from sklearn.ensemble import RandomForestRegressor
X = data['number_of_sessions', 'number_of_rounds', 
	'number_of_days', 'number_of_purchases', 
	'total_purchase_amount']
y = data["day_360_CLV"]
model = RandomForestRegressor(n_estimators=100)
model.fit(X, y)
\end{lstlisting}

For readability, the number of features selected as inputs is reduced in comparison to~\cite{Sifa2018}; furthermore, the number of trees in the ensemble has been chosen without any specific motivation as no precise description is included in the article. Both the size of the ensemble and maximum depth the trees, as well as the other parameters of the ensemble, are problem specific and should be selected either algorithmically or through extensive experimentation. 

The code necessary to perform CLV prediction using the previously fit model looks as follows:
\begin{lstlisting}[language=Python, backgroundcolor=\color{lightgray}, frame=line]
X = new_data['number_of_sessions', 'number_of_rounds', 
	'number_of_days', 'number_of_purchases', 
	'total_purchase_amount']
y = model.predict(X)
\end{lstlisting}

The resulting array $y$ will contain the predicted customer lifetime values. The sci-kit learn library includes also a number of functionalities to estimate the predictions error and to perform different forms of validation and testing -- e.g. k-folds cross-validation. More information can be found on the project's documentation page\footnote{https://scikit-learn.org/stable/documentation.html}.

\section{Conclusions and Future Directions} \label{sec:future}

In this chapter we have described the applications of customer lifetime value prediction in the free-to-play games and other industries and we attempted to give a comprehensive description of the different methods that are currently used to predict CLV. 
We have identified a number of activities that can benefit from an accurate prediction ranging from customer acquisition to market segmentation and we have described how the need for such prediction is even more important in free-to-play games due to the characteristics of the marker and the customer relationship -- e.g. incredibly high competition in user acquisition and very low number of paying customers compared to the player base.

In Section \ref{sec:models}, we have identifies within the literature four main group of methods for the prediction: average based, Pareto/NBD and derivatives, Markov Chain Models and Supervised Learning Models. The methods that are currently dominant within the free-to-play games industry are either average based or some form of Pareto/NBD. However, we can see an emerging trend in recent years of more studies being published on the application of different supervised learning algorithms to CLV prediction. 

These methods have a number of advantages over classical statistical methods in this context as they make no assumption of the distributions of the input data and easily allow multiple co-variates to be included in the model. 
Furthermore, in an industry with such a low rate of conversion to paying users, being able to predict revenue from a player that has not yet made any purchase has the potential to be very useful, for instance, for early estimation of the profitability of a player acquisition campaign.

Based on the analysis presented and in light of the recent improvements within the field of machine learning, we close the article by proposing a number of possible interesting future directions for customer lifetime value prediction research.

\paragraph{Deep Learning} One group of supervised learning algorithms that has only minimally been explored in one of the most recent articles~\cite{Sifa2018} is Deep Neural Networks. The work by Sifa et al. don't give an in-depth description of the deep multi-layer perceptron used; however, in the absence of other information, we can assume that the model was based on a standard fully-connected feed-forward neural network with multiple hidden layers. Given the results achieved with a relatively simple architecture, it would be interesting to test more advanced deep learning techniques such as auto-encoders for feature extraction~\cite{Vincent2008} or deep convolutional neural networks~\cite{krizhevsky2012nips}.

\paragraph{Time Series} All of the models presented in this chapter act on some form of summary data representing the behavior of customers/players up to a given point in time. While some of these summary representations do include some representation of the temporal aspect of the behaviors -- e.g. the recency and frequency features in RFM based models -- these cannot capture any particular sequences of purchases or any sequence of in-game events that is connected to the probability that a player will perform a purchase or not.
One possible approach to leverage this kind of information is to treat player behavior data as time series and perform some form of time-series regression or classification to predict CLV or a purchase event. Within aforementioned deep learning field algorithms such as Long-short term memory networks~\cite{Hochreiter1997} or the more recent temporal convolution networks~\cite{Elbayad2018}.

\paragraph{Transfer Learning and Lifelong Learning} Game developers in the mobile free-to-play market are challenged with the increasing need to manage a multitude of games live at the same time. These games have often very long lifetimes and, throughout their lifetimes, the game is adjusted and evolved and the player base changes dues to the game evolution and the changes in the customer acquisition initiatives.
This adds a new dimension of the problem of CLV prediction as models need to be constantly updated and new models need to be built quickly for new version of the games and for new games. 
Within this scenario, research in cross-game player behavior analysis~\cite{Martinez2011acii}, transfer learning~\cite{Yosinski2014a} and life-long machine learning~\cite{Silver2013}, and their application to CLV prediction will become increasingly important.

\bibliographystyle{plain}
\bibliography{library}

\end{document}